\documentclass[11pt,a4paper]{article}
\usepackage[hyperref]{emnlp2020}
\usepackage{times}
\usepackage{latexsym}

\usepackage{microtype}

\aclfinalcopy 

\title{SacreROUGE: An Open-Source Library for Using and Developing Summarization Evaluation Metrics}

\author{
  Daniel Deutsch and Dan Roth \\
  Department of Computer and Information Science \\
  University of Pennsylvania \\
  \texttt{\{ddeutsch,danroth\}@seas.upenn.edu}
}

\date{}

\begin{document}
\maketitle

\begin{abstract}
We present SacreROUGE, an open-source library for using and developing summarization evaluation metrics.\footnote{\url{https://github.com/danieldeutsch/sacrerouge}}
SacreROUGE removes many obstacles that researchers face when using or developing metrics:
(1) The library provides Python wrappers around the official implementations of existing evaluation metrics so they share a common, easy-to-use interface;
(2) it provides functionality to evaluate how well any metric implemented in the library correlates to human-annotated judgments, so no additional code needs to be written for a new evaluation metric; 
and (3) it includes scripts for loading datasets that contain human judgments so they can easily be used for evaluation.
This work describes the design of the library, including the core \texttt{Metric} interface, the command-line API for evaluating summarization models and metrics, and the scripts to load and reformat publicly available datasets.
The development of SacreROUGE is ongoing and open to contributions from the community.
\end{abstract}
\section{Introduction}
Evaluating models is a critical step of the machine learning workflow.
However, unlike classification-based tasks, evaluating models which generate text is difficult and is a research area on its own.
The basic workflow for developing a new automatic evaluation metric is to design/implement the metric, calculate its correlation to human judgments, then use that metric to evaluate text generation systems.

While there have been significant efforts to build libraries for developing machine learning models \citep{Klein2017,Gardner2018,Ott2019}, no equivalent library exists for developing evaluation metrics.
In this work, we present SacreROUGE, an open-source, Python-based library for using and developing text generation metrics, with an emphasis on summarization.

SacreROUGE removes many obstacles that researchers face when they use or develop evaluation metrics.
First, the official implementations of various metrics do not share a common interface or programming language, so using many metrics to evaluate a model can be frustrating and time consuming.
SacreROUGE provides Python-based wrappers around many evaluation metrics so they all implement a simple, easy-to-use interface regardless of how they are implemented internally (\S\ref{sec:metric-interface}).

Second, evaluating metrics themselves can be tricky.
Correlations between metric values and human judgments are calculated in several different ways, there are multiple commonly used correlation coefficients, and fairly comparing human-written references to system output requires implementing jackknifing.
Since the evaluation code in SacreROUGE is shared across all of the metrics, any metric with implements the common \texttt{Metric} interface can be evaluated without writing additional code (\S\ref{sec:evaluation}).

Third, datasets that contain judgments which are commonly used to evaluate metrics do not share the same format, so writing code to load each dataset requires writing a significant amount of effort.
SacreROUGE provides scripts for popular summarization datasets that load and reformat them into a common schema so they can easily be used for evaluation (\S\ref{sec:datasets}).

The development of SacreROUGE is ongoing.
We intend to add more metrics and datasets to the library as they become available.
Further, we encourage researchers to use the SacreROUGE framework to use existing metrics and develop new ones.
SacreROUGE is released under the Apache 2.0 license and is open to contributions from the community.

\section{The Metric Interface}
\label{sec:metric-interface}
The development of evaluation metrics for summarization has been an active area of research for two decades.
However, the community has not converged on a consistent format for the input data, so each metric uses its own custom schema.
Further, the published code for evaluation metrics is written in various programming languages based on which language was popular when the metric was proposed.
These challenges make it very cumbersome to use multiple metrics to evaluate a summarization system.
SacreROUGE addresses these two problems by unifying all of the metrics' implementations into a common interface called \texttt{Metric}.
The interface provides a Pythonic API that allows for evaluating an individual summary or batch of summaries.
Since all of the metrics share the same interface, evaluating a summarization system with several different metrics is trivial.

In order to support older evaluation metrics written in languages such as Perl or Java, we have written Python wrappers around the original code that still implement the \texttt{Metric} interface.
Internally, the wrappers serialize the input summaries to the format required by the underlying metric, a subprocess is created to run the original metric's code, and the output is then loaded from disk again in Python.
This way, we do not have to port the original metric's code to Python and end-users can still use the metrics with the Python API.

SacreROUGE currently supports the following evaluation metrics:
AutoSummENG \citep{Giannakopoulos2008},
BERTScore \citep{Zhang2019},
BEwT-E \citep{Tratz2008},
METEOR \citep{Denkowski2014},
MeMoG \citep{Giannakopoulos2010},
NPowER \citep{Giannakopoulos2013},
ROUGE \citep{Lin2004},
a near-identical Python-based version of ROUGE that we wrote,
SIMetrix \citep{Louis2009},
and SumQE \citep{Xenouleas2019}.

\paragraph{Handling Dependencies}
Many of the evaluation metrics rely on external resources in the form of code, models, or data files.
Setting up these dependencies in the right format to use the metrics can be difficult.

The SacreROUGE library addresses this problem by providing setup scripts for each metric which download or compile any required resources.
To make this process as easy as possible for the end-user, these scripts are run through a \texttt{setup-metric} command.
The command takes the name of the metric to setup, then downloads the required dependencies to a common folder which is managed by SacreROUGE.
Abstracting the metric setup by a simple command makes it such that the end-user can quickly and easily begin using all of the metrics within the library.
\section{Evaluating Systems and Metrics}
\label{sec:evaluation}
The two most common use cases of an evaluation metric are to evaluate a summarization system and to evaluate a metric itself by calculating its correlation to human judgments.
Since all of the metrics in SacreROUGE implement a common interface, the code for these procedures is shared, so developers of new metrics do not need to rewrite the code to implement these procedures.
This logic is exposed through \texttt{evaluate}, \texttt{score}, and \texttt{correlate}, which are subcommands of \texttt{sacrerouge}, the entry point for the library's command-line interface.

\paragraph{The \texttt{evaluate} Subcommand}
The \texttt{evaluate} subcommand accepts a specific metric and an input file that contains the output of a summarization system for an input corpus.
The command will load the input data, pass it to the metric, and save the metric's output at the summary-level and system-level.
The summary-level output contains the metric's value for each individual summary, whereas system-level output represents the average performance across the dataset and is most often reported in papers.

\paragraph{The \texttt{score} Subcommand}
Evaluating metrics themselves is exposed through the \texttt{score} and \texttt{correlate} subcommands.
The \texttt{score} subcommand is very similar to \texttt{evaluate} except for two key differences.
First, the input data is not expected to the the output from a single system;
Correlations to human judgment often involve scoring summaries from many different summarization models.
Subsequently, no system-level metrics are calculated.

Second, the \texttt{score} subcommand will run jackknifing on the input data when possible and necessary.
Jackknifing is a procedure which allows the value of a metric on system-produced and human-written summaries to be fairly compared when the human-written summaries are used to assess the quality of the system summary.
Briefly, if there is more than one reference summary, each reference is evaluated against all of the others.
Each system summary is repeatedly evaluated against each possible subset of the reference summaries that has one reference removed.
The final system summary score is an average across those evaluations.
When jackknifing is performed, a \texttt{\_jk} suffix is appended to the name of the metric which makes it clear that it is not comparable to the non-jackknifed version.

\paragraph{The \texttt{correlate} Subommand}
After the \texttt{score} subcommand is complete, the \texttt{correlate} subcommand can be used to calculate the correlation between two metrics.
SacreROUGE calculates the three correlation coefficients most commonly used in summarization: Pearson, Spearman, and Kendall.
Further, these correlations are computed at three different granularities: the summary-level, the system-level, and globally.
The summary-level correlation calculates the average correlation per input.
The system-level calculates the correlation between average system performances for each metric.
The global correlation directly calculates the correlation between all of the observed metric values.
The former two granularities are most often used in the summarization literature.

\paragraph{Handling Different Input Requirements}
It is often the case that different metrics require different input data (e.g., some metrics use reference summaries, others need access to the input documents).
Therefore, the required data must be loaded from the input file and the \texttt{evaluate} and \texttt{score} subcommands must pass the required data to the metric.

The interface for loading data from an input file in SacreROUGE is called a \texttt{DatasetReader}.
For a given input file(s), a \texttt{DatasetReader} loads the \texttt{Field}s for the evaluation instances.
A \texttt{Field} is a base class which contains the data for an input instance, such as a \texttt{DocumentsField} that maintains the contents of the input documents.
Then, each evaluation instance contains a mapping from the name of a field to its data.

In order to pass the appropriate \texttt{Field}s to the summarization metrics, we require that every class that implements the \texttt{Metric} interface lists the names of the \texttt{Field}s that it uses.
For instance, the wrapper for the document-based evaluation metric SIMetrix specifies it needs a field called \texttt{documents}, a key in the evaluation instance \texttt{Field} mapping.
Then, once the input data has been loaded, the \texttt{evaluate} and \texttt{score} commands can pass the required data to a metric for evaluation.

\paragraph{Automatically Generated Subcommands}
It is desirable to have a different \texttt{evaluate} and \texttt{score} subcommand for each individual metric so that developers can easily specify different metric parameters on the command line.
A naive implementation of this would require manually creating the subcommand for each metric.
However, in order to eliminate as much boilerplate code as possible, SacreROUGE includes a feature to automatically generate these subcommands for any metric that implements the \texttt{Metric} interface.

Using Python's \texttt{inspect} and \texttt{typing} libraries, we are able to examine the constructor of each metric and generate a command-line argument for each parameter.
For parameters with primitive types, the \texttt{argparse} library directly supports casting command line parameters to the correct types.
However, some metrics may use complex types, such as a list of integers.
In such situations, SacreROUGE assumes that the command line argument will be a string-serialized JSON object that can be deserialized into the required type at runtime.
This allows us to support automatically generating \texttt{evaluate} and \texttt{score} subcommands for every metric supported by the library.

\section{A Common Dataset Format}
\label{sec:datasets}
Over the past two decades, the summarization community has collected a large number of expensive summarization dataset and human quality annotations.
However, these very useful datasets are seldom saved in a common format, forcing every researcher who wants to train a model on the datasets or use the judgments to evaluate a metric to write boilerplate code to load the data.

To mitigate this issue, SacreROUGE provides scripts that will load the datasets and their corresponding judgments, then serialize them to new files with a common format.
The data is serialized in such a manner that it can be directly used in the \texttt{evaluate}, \texttt{score}, and \texttt{correlate} subcommands, thereby making it incredibly easy to run or evaluate any metric in the library on the dataset.

The scripts to preprocess the datasets are exposed through the \texttt{setup-dataset} subcommand.
The subcommand accepts the name of a dataset, an output directory, and any potential dataset-specific arguments.
Then, SacreROUGE will load and preprocess the respective dataset.
For datasets which are publicly available, the scripts will download the data automatically.
However, many summarization datasets are licensed, so the corresponding preprocessing scripts require paths to the original data supplied to the command.

The datasets which are currently supported by SacreROUGE are the Document Understanding Conference from 2001 to 2007,\footnote{\url{https://duc.nist.gov/}} Text Analysis Conference from 2008 to 2011,\footnote{\url{https://tac.nist.gov/}} the MultiLing 2011, 2013, 2015, 2017, and 2019 Workshops,\footnote{\url{http://multiling.iit.demokritos.gr/}} and the CNN/DailyMail dataset judgments provided by \citet{Chaganty2018}.
We intend to add more datasets as they become available.
\section{Related Work}
The namesake and idea for SacreROUGE came from the SacreBLEU \citep{Post2018} library.
SacreBLEU was developed to standardize and simplify calculating BLEU \citep{Papineni2002} for machine translation.
Like SacreROGUE, it provides a simple command-line interface to download and evaluate on common machine translation datasets.
Whereas SacreBLEU is mainly for evaluating machine translation models with BLEU, our library focuses on summarization and includes a large number of evaluation metrics.
Further, SacreROUGE also provides a framework for developing and evaluating new metrics.

Much of the design of SacreROUGE was inspired by AllenNLP \citep{Gardner2018}, a library built on PyTorch \citep{Paszke2017} for developing deep learning models.
AllenNLP provides useful abstractions over different models and neural network modules that allows for the sharing of boilerplate code so developers can quickly create and train new machine learning models.
SacreROUGE provides similar abstractions for evaluation metrics.

Recently, Hugging Face released a library called \texttt{nlp} that sets out to achieve similar goals to SacreROUGE.\footnote{\url{https://github.com/huggingface/nlp}}
Namely, they also standardize loading different datasets and provide a Pythonic API to many popular evaluation metrics.
However, because their library is focused on a large number of NLP tasks and SacreROUGE is built specifically for summarization, SacreROUGE is able to support more summarization datasets and metrics.
Further, unlike SacreROUGE, \texttt{nlp} does not provide a framework for developing and evaluating new metrics.
\section{Conclusion}
We have presented SacreROUGE, an open-source library dedicated to the development of summarization evaluation metrics.
With a unified metric interface and common data format, our library makes it very simple to use existing evaluation metrics as well as develop new ones with a minimum amount of effort.
We hope that future researchers will contribute their own metrics and datasets to the library so that it is as easy as possible to run and evaluate summarization metrics.

\bibliographystyle{acl_natbib}
\bibliography{summarization,emnlp2020}

\end{document}